\documentclass[10pt,twocolumn,letterpaper]{article}

\usepackage[pagenumbers]{cvpr} % To force page numbers, e.g. for an arXiv version
\usepackage{graphicx}
\usepackage{amsmath}
\usepackage{amssymb}
\usepackage{booktabs}
\usepackage{multirow}
\usepackage{enumerate}
\usepackage[pagebackref,breaklinks,colorlinks]{hyperref}
\makeatletter
\def\@fnsymbol#1{\ensuremath{\ifcase#1\or \dagger\or \ddagger\or
   \mathsection\or \mathparagraph\or \|\or **\or \dagger\dagger
   \or \ddagger\ddagger \else\@ctrerr\fi}}
\makeatother

\usepackage[capitalize]{cleveref}
\crefname{section}{Sec.}{Secs.}
\Crefname{section}{Section}{Sections}
\Crefname{table}{Table}{Tables}
\crefname{table}{Tab.}{Tabs.}

\def\cvprPaperID{6314} % *** Enter the CVPR Paper ID here
\def\confName{CVPR}
\def\confYear{2023}

\begin{document}

%%%%%%%%% TITLE - PLEASE UPDATE
\title{Semantic Ray: Learning a Generalizable Semantic Field with Cross-Reprojection Attention}

% \author{First Author\\
% Institution1\\
% Institution1 address\\
% {\tt\small firstauthor@i1.org}
% % For a paper whose authors are all at the same institution,
% % omit the following lines up until the closing ``}''.
% % Additional authors and addresses can be added with ``\and'',
% % just like the second author.
% % To save space, use either the email address or home page, not both
% \and
% Second Author\\
% Institution2\\
% First line of institution2 address\\
% {\tt\small secondauthor@i2.org}
% }
\author{Fangfu Liu$^{1,3}$, Chubin Zhang$^{2}$, Yu Zheng$^{2}$, Yueqi Duan$^{1\dagger}$\\
$^{1}$Department of Electronic Engineering, Tsinghua University\\
$^{2}$Department of Automation, Tsinghua University\\
$^{3}$Beijing National Research Center for Information Science and Technology\\
{\tt\small \{liuff19,zhangcb19,zhengyu19\}@mails.tsinghua.edu.cn, }
{\tt\small duanyueqi@tsinghua.edu.cn}}

\makeatletter
\let\@oldmaketitle\@maketitle% Store \@maketitle
\renewcommand{\@maketitle}{\@oldmaketitle
\begin{minipage}{\textwidth}
\centering
\vspace{-0.8cm}
\includegraphics[width=1\linewidth]{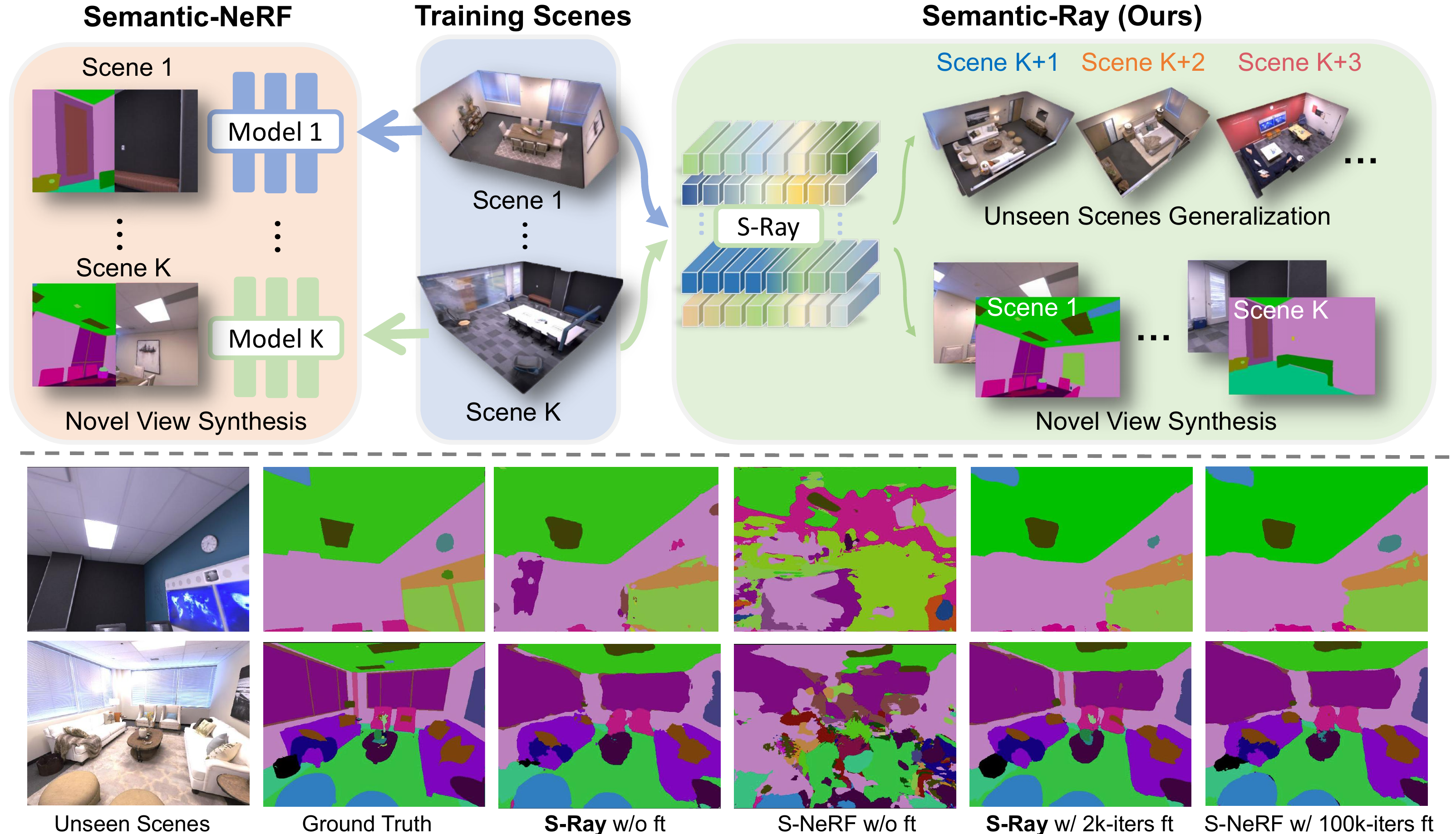}
\vspace{-6mm}
\captionof{figure}{\textbf{Top}: Comparisons between Semantic-NeRF~\cite{semantic-nerf} and our method Semantic-Ray. Semantic-NeRF (S-NeRF for short) needs to train one specific model for each scene, while our Semantic-Ray (S-Ray for short) trains one unified model on multiple scenes and generalizes to unseen scenes. \textbf{Bottom}: Experimental comparisons between S-Ray and S-NeRF on generalization ability. We observe that our network S-Ray can effectively \textit{fast generalize} across diverse unseen scenes while S-NeRF fails in a new scene. Moreover, our result can be improved by fine-tuning on more images for only 10 min ($2k$ iterations), which achieves comparable quality with the Semantic-NeRF's result for $100k$ iterations per-scene optimization.}
\vspace{-0.2cm}
\label{fig:teaser}
\end{minipage}
\bigskip
}% ... an image

\maketitle

\newcommand\blfootnote[1]{% 
\begingroup 
\renewcommand\thefootnote{}\footnote{#1}% 
\addtocounter{footnote}{-1}% 
\endgroup 
}
\blfootnote{\textsuperscript{\dag}Corresponding author.}

\begin{abstract}
In this paper, we aim to learn a semantic radiance field from multiple scenes that is accurate, efficient and generalizable. While most existing NeRFs target at the tasks of neural scene rendering, image synthesis and multi-view reconstruction, there are a few attempts such as Semantic-NeRF that explore to learn high-level semantic understanding with the NeRF structure. However, Semantic-NeRF simultaneously learns color and semantic label from a single ray with multiple heads, where the single ray fails to provide rich semantic information. As a result, Semantic NeRF relies on positional encoding and needs to train one specific model for each scene. To address this, we propose Semantic Ray (S-Ray) to fully exploit semantic information along the ray direction from its multi-view reprojections. As directly performing dense attention over multi-view reprojected rays would suffer from heavy computational cost, we design a Cross-Reprojection Attention module with consecutive intra-view radial and cross-view sparse attentions, which decomposes contextual information along reprojected rays and cross multiple views and then collects dense connections by stacking the modules. Experiments show that our S-Ray is able to learn from multiple scenes, and it presents strong generalization ability to adapt to unseen scenes. Project page: \href{https://liuff19.github.io/S-Ray/}{https://liuff19.github.io/S-Ray/}.
\end{abstract}
%%%%%%%%% BODY TEXT
% \input{cvpr2023/Chapters/1-introduction}
\section{Introduction}

Recently, Neural Radiance Field (NeRF) \cite{NeRF}, a new novel view synthesis method with implicit representation, has taken the field of computer vision by storm \cite{NeRF_review2022}. NeRF and its variants \cite{NeRF,nerf++, mipnerf, pixelnerf} adopt multi-layer perceptrons (MLPs) to learn continuous 3D representations and utilize multi-view images to render unseen views with fine-grained details. NeRF has shown state-of-the-art visual quality, produced impressive demonstrations, and inspired many subsequent works~\cite{mvsNeRF, GeoNeRF, code_nerf, IBRnet, pixelnerf}.

While the conventional NeRFs have achieved great success in low- and middle-level vision tasks such as neural scene rendering, image synthesis, and multi-view reconstruction \cite{mvsNeRF, neuray, kilo_nerf, Fast_nerf, grf, GIRAFFE, UNISURF}, it is interesting to explore their more possibilities in high-level vision tasks and applications. Learning high-level semantic information from 3D scenes is a fundamental task of computer vision with a wide range of applications \cite{dl_for_medical_img_seg, autonomous_driving_seg, garcia2017review, robotic_semseg}. For example, a comprehensive semantic understanding of scenes enables intelligent agents to plan context-sensitive actions in their environments. One notable attempt to learn interpretable semantic understanding with the NeRF structure is Semantic-NeRF \cite{semantic-nerf}, which regresses a 3D-point semantic class together with radiance and density. Semantic-NeRF shows the potential of NeRF in various high-level tasks, such as scene-labeling and novel semantic view synthesis.

However, Semantic-NeRF follows the vanilla NeRF by estimating the semantic label from a single ray with a new semantic head. While this operation is reasonable to learn low-level information including color and density, a single ray fails to provide rich semantic patterns -- we can tell the color from observing a single pixel, but not its semantic label. To deal with this, Semantic-NeRF heavily relies on positional encoding to learn semantic features, which is prone to overfit the current scene and only applicable to novel \textit{views} within the same scene~\cite{NeSF}. As a result, Semantic-NeRF has to train one model from scratch for every scene independently or provides very limited novel scene generalization by utilizing other pretrained models to infer 2D segmentation maps as training signals for unseen scenes. This significantly limits the range of applications in real-world scenarios.

In this paper, we propose a neural semantic representation called \textbf{Semantic Ray} (S-Ray) to build a generalizable semantic field, which is able to learn from multiple scenes and directly infer semantics on novel viewpoints across novel scenes as shown in Figure~\ref{fig:teaser}. 
% To our best knowledge, this is the first work to learn a generalizable semantic field in real-world scenes. 
As each view provides specific high-level information for each ray regarding of viewpoints, occlusions, etc., we design a Cross-Reprojection Attention module in S-Ray to fully exploit semantic information from the reprojections on multiple views, so that the learned semantic features have stronger discriminative power and generalization ability. While directly performing dense attention over the sampled points on each reprojected ray of multiple views would suffer from heavy computational costs, we decompose the dense attention into intra-view radial and cross-view sparse attentions to learn comprehensive relations in an efficient manner.

More specifically, for each query point in a novel view, different from Semantic-NeRF that directly estimates its semantic label with MLP, we reproject it to multiple known views. It is worth noting that since the emitted ray from the query point is virtual, we cannot obtain the exact reprojected point on each view, but a reprojected ray that shows possible positions. Therefore, our network is required to simultaneously model the uncertainty of reprojection within each view, and comprehensively exploit semantic context from multiple views with their respective significance. To this end, our Cross-Reprojection Attention consists of an intra-view radial attention module that learns the relations among sampled points from the query ray, and a cross-view sparse attention module that distinguishes the same point in different viewpoints and scores the semantic contribution of each view. As a result, our S-Ray is aware of the scene prior with rich patterns and generalizes well to novel scenes. We evaluate our method quantitatively and qualitatively on synthetic scenes from the Replica dataset \cite{replica} and real-world scenes from the ScanNet dataset \cite{scannet}. Experiments show that our S-Ray successfully learns from multiple scenes and generalizes to unseen scenes. By following Semantic-NeRF \cite{semantic-nerf}, we design competitive baselines based on the recent MVSNeRF \cite{mvsNeRF} and NeuRay \cite{neuray} architectures for generalizable semantic field learning. Our S-Ray significantly outperforms these baselines which demonstrates the effectiveness of our cross-reprojection attention module.

% 如果实验结果有时间优势，一定放到这个里面！！！！！

\begin{figure*}[!t]
    \centering
    \includegraphics[width=\linewidth]{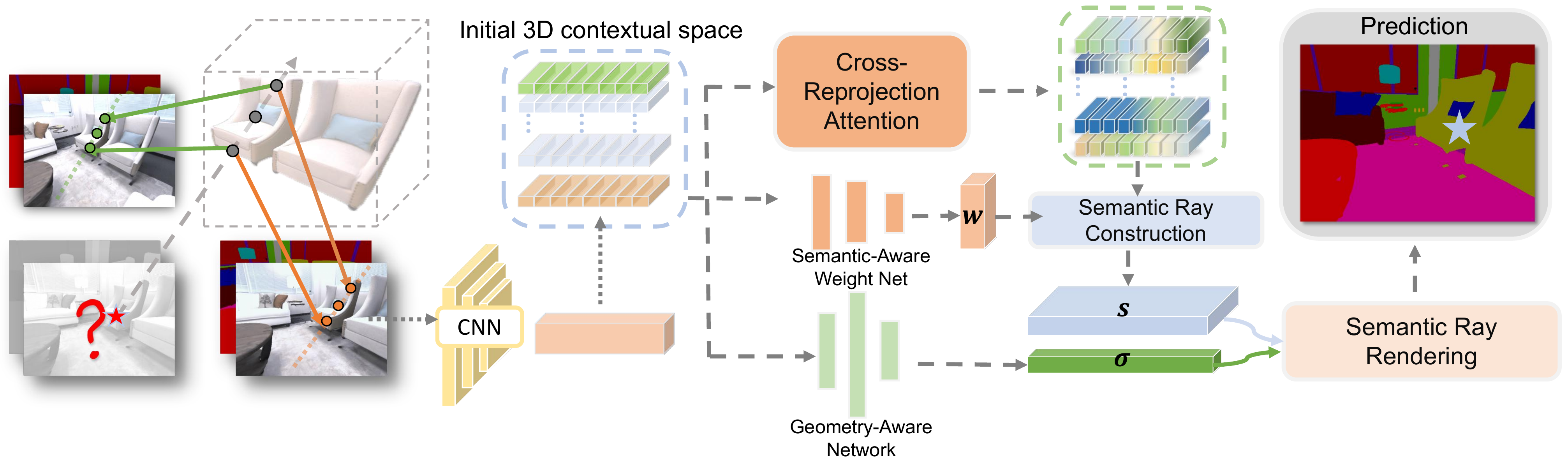}
    \vspace{-7mm}
    \caption{\textbf{Pipeline of semantic rendering with S-Ray}. Given input views and a query ray, we first reproject the ray to each input view and apply a CNN-based module to extract contextual features to build an initial 3D contextual space (Sec.~\ref{sec:contextualSpace}). Then, we apply the Cross-Reprojection Attention module to learn dense semantic connections and build a refined contextual space (Sec.~\ref{sec:cross-reprojection}). For semantic ray rendering, we adopt the semantic-aware weight net to learn the significance of each view to construct our semantic ray from refined contextual space (Sec.~\ref{sec:semantic-ray}). Finally, we leverage the geometry-aware net to get the density and render the semantics along the query ray.}
    \label{fig:pipeline}
    \vspace{-10pt}
\end{figure*}
\section{Related Work}
\noindent \textbf{Semantic Segmentation.}
Semantic segmentation is one of the high-level tasks that paves the way toward complete scene understanding, with most methods targeting a fully supervised, single-modality problem (2D \cite{SegNet, DeepLab, rethinking_atrous_for_seg, FCN, U-net} or 3D \cite{segsota-review, cylindrical_conv_seg, survey_3d_point_cloud, rangenet++}). Recently, machine learning methods have proven to be valuable in semantic segmentation \cite{U-net, criss-cross-attention, SegNet, SegGroup} which aims to assign a separate class label to each pixel of an image. However, most methods suffer from severe performance degradation when the scenes observed at test time mismatch the distribution of the training data \cite{genralizable_model_agnostic_semseg, semseg_ood}. To alleviate the issue, 2D-based architectures \cite{U-net, FCN, garcia2017review} are often trained on large collections of costly annotated data \cite{COCO_dataset} while most 3D prior works \cite{lidar_seg, kinectfusion, Dense_vis_tracking, JSENet, pointnet,pointnet++} rely on 3D sensors. Though straightforward to apply, 2D-based approaches only produce per-pixel annotations and fail to understand the 3D structure of scenes \cite{NeSF} and 3D sensors are too expensive to be widely available as RGB cameras. In contrast to these methods, our method reconstructs and then segments a 3D semantic representation from 2D inputs and supervision alone without ground truth 3D annotations or input geometry.

% \noindent \textbf{Implicit 3D Representations.} The implicit 3D representations have recently emerged as a promising direction to recover 3D geometries. It is initially formulated as level sets by optimizing neural nets that map the coordinates to an occupancy field \cite{Occupancy_networks}. Some methods use implicit representations to reconstruct a 3D scene \cite{conv_occupancy_net, iMAP} or shape \cite{local_sdf, bspnet, learning_implicit_field_gen, neural_geo_level_of_detail}. Similar to our target, Atlas~\cite{atlas} learns a 3D implicit TSDF reconstruction from 2D images and can also segment the predicted scene geometry. However, this method requires 3D data supervision, while our approach uses only 2D images at both training and test time.

\noindent \textbf{Neural Radiance Fields.} 
Recently, implicit neural representations have advanced neural processing for 3D data and multi-view 2D images \cite{SRN, Occupancy_networks, Deep_SDF, multiview_by_disentangle}. In particular, Neural Radiance Fields (NeRF) \cite{NeRF} has drawn great attention, which is a fully-connected neural network that can generate novel views of complex 3D scenes, based on a partial set of 2D images. A NeRF network aims to map from 3D location and viewing direction (5D input) to opacity and color (4D input). Several following works emerge trying to address its limitations and improve its performance, including fast training \cite{direct_voxel_grid_opt_super_fast, Depth-supervised_NeRF}, efficient inference \cite{DeRF, Neural_sparse_voxel_field, Fast_nerf, kilo_nerf, PlenOctrees}, unbounded scenes training \cite{nerf++, mipnerf}, better generalization \cite{GRAF, grf, IBRnet, pixelnerf, GeoNeRF, GPNR}, generative modeling \cite{GNeRF, GIRAFFE, GRAF}, editing \cite{editing_conditional_RF, clip-nerf, code_nerf}. 
As NeRF achieves very impressive results for novel view synthesis, researchers start to explore high-level tasks in NeRF such as semantic segmentation \cite{semantic-nerf, NeSF}. However, Semantic-NeRF\cite{semantic-nerf} is only applicable in the single-scene setting while NeSF \cite{NeSF} only conducts experiments in synthetic data with insufficient generalization and high computational costs. In contrast, taking NeRF as a powerful implicit scene representation, our method can learn a generalizable semantic representation of new, real-world scenes with high quality. 
% if add the however to claim the current segmentation task

\section{Method}

Given a set of $N$ input views with known camera poses, our goal is to synthesize novel semantic views from arbitrary angles across unseen scenes with strong generalizability. Our method can be divided into three stages: (1) build the 3D contextual space from source multiple views, (2) decompose the semantic information from 3D space along reprojected rays and cross multiple views with cross-reprojection attention, (3) reassemble the decomposed contextual information to collect dense semantic connections in 3D space and re-render the generalizable semantic field. Our pipeline is depicted in Figure~\ref{fig:pipeline}. Before introducing our S-Ray in detail, we first review the volume rendering on a radiance field \cite{volume_rendering_origin, NeRF}.
\subsection{Preliminaries}
\noindent \textbf{Neural Volume Rendering.} Neural volume rendering aims to learn two functions: $\sigma(\mathbf{x}; \theta): \mathbb{R}^3 \mapsto \mathbb{R}$, which maps the spatial coordinate $\mathbf{x}$ to a density $\sigma$, and $\mathbf{c}(\mathbf{x}, \mathbf{d}; \theta): \mathbb{R} ^3 \times \mathbb{R}^3 \mapsto \mathbb{R}^3$ that maps a point with the viewing direction to the RGB color $\mathbf{c}$. The density and radiance functions are defined by the parameters $\theta$. To learn these functions, they are evaluated through a ray emitted from a query view. The ray is parameterized by $\mathbf{r}(t)=\mathbf{o}+t \mathbf{d}$, $t \in\left[t_n, t_f\right]$, where $\mathbf{o}$ is the start point at the camera center, $\mathbf{d}$ is the unit direction vector of the ray, and $\left[t_n, t_f\right]$ is the near-far bound along the ray. Then, the color for the associated pixel of this ray can be computed through volume rendering \cite{volume_rendering_origin}:
\begin{align}
    \hat{\mathbf{C}}(\mathbf{r} ; \theta)=\int_{t_n}^{t_f} T(t) \sigma(\mathbf{r}(t)) \mathbf{c}(\mathbf{r}(t), \mathbf{d}) \mathrm{d} t, 
\end{align}
where $T(t)=\exp \left(-\int_{t_n}^t \sigma(\mathbf{r}(s)) \mathrm{d} s\right)$. In practice, the continuous integration can be approximated by a summation of discrete samples along the ray by the quadrature rule. For selected $N$ random quadrature points $\{t_k\}_{k=1}^N$ between $t_n$ and $t_f$, the approximated expected color is computed by:
\begin{align}
\label{eq:render_discrete}
    \hat{\mathbf{C}}(\mathbf{r}; \theta) = \sum_{k=1}^{N} T(t_k)\alpha(\sigma(t_k)\delta_k) \mathbf{c}(t_k), 
\end{align}
where $T(t_k) = \exp \left(-\sum_{k^{'}=1}^{k-1} \sigma\left(t_k\right) \delta_k\right)$, $\alpha(x) = 1 - exp(-x)$, and $\delta_k = t_{k+1} - t_k$ are intervals between sampled points. 

\subsection{Building 3D Contextual Space across Views}
\label{sec:contextualSpace}
% if add related: In contrast to IBRnet: only reproject a point, we reproject a ray
NeRF-based generalization rendering methods \cite{grf, pixelnerf, IBRnet} construct a radiance field by reprojecting a single point of  the query ray to source views and extracting the point-based feature. It is reasonable to predict color from a single point but is quite insufficient for semantics which needs more contextual information. Therefore, we build the 3D contextual space to learn rich semantic patterns by reprojecting the whole query ray across views. Given the $N$ points $\{\mathbf{p}_{i}\}_{i = 1,2,..., N}$ sampled from the ray emitting from the query view and known camera pose (\ie, the rotation matrix $\mathbf{R}$ and the translation vector $\mathbf{t}$). Without losing generality, we can rewrite the ray as:
\begin{equation}
    \mathbf{r}(z) = \mathbf{p}_i + z\frac{\mathbf{p}_j - \mathbf{p}_i}{||\mathbf{p}_j - \mathbf{p}_i||}, \quad z\in \mathbb{R}.
\end{equation}
Then, the ray warping function is defined as:
\begin{equation}
    w(\mathbf{r}(z), \mathbf{R}, \mathbf{t}) := \mathbf{K} \pi (\mathbf{R} \cdot \mathbf{r}(z) + \mathbf{t}), 
\end{equation}
which allows reprojecting the ray onto source views to obtain the plane ray $\mathbf{r}^{*}(z)$, \ie, $ \mathbf{r}^{*}(z) = w(\mathbf{r}(z), \mathbf{R}, \mathbf{t})$, where $\mathbf{K}$ is the camera calibration matrix and $\pi (\mathbf{u}):= [\mathbf{u}_x / \mathbf{u}_z, \mathbf{u}_y / \mathbf{u}_z]$ is the projection function. 

Let $\mathcal{F}_j(\mathbf{r}^{*}(z)) (j = 1, 2, ..., m)$ denote the ray-based feature of the query ray reprojected on the $j$-th source view,  $\mathcal{F}_j(\mathbf{r}^{*}(z_i))$ be the $\mathbf{p}_i$ point-based feature within the $j$-th source view. 
Due to the permutation invariance of the source views, we use a shared U-Net-based convolutional neural network to extract dense contextual information $\mathcal{F}$ from these views. Then, we build our 3D contextual space $\mathcal{M}$ across views as:

\small
\begin{equation}
   \mathcal{M}
=\begin{bmatrix}
\mathcal{F}_1(\mathbf{r}^{*}(z_1)) &  \mathcal{F}_1(\mathbf{r}^{*}(z_2))  & \cdots\ &\mathcal{F}_1(\mathbf{r}^{*}(z_N))\\
\mathcal{F}_2(\mathbf{r}^{*}(z_1)) &  \mathcal{F}_2(\mathbf{r}^{*}(z_2)) & \cdots\ & \mathcal{F}_2(\mathbf{r}^{*}(z_N))\\
 \vdots   & \vdots & \ddots  & \vdots  \\
 \mathcal{F}_m(\mathbf{r}^{*}(z_1)) & \mathcal{F}_m(\mathbf{r}^{*}(z_2)) & \cdots\ & \mathcal{F}_m(\mathbf{r}^{*}(z_N))\\
\end{bmatrix}
\end{equation}
\normalsize
which is a 3D matrix (\ie, $\mathcal{M} \in \mathbb{R}^{m \times N \times C}$) to describe a full space contextual information around the query ray with feature dimension $C$.

\begin{figure*}
    \centering
    \includegraphics[width=1\linewidth]{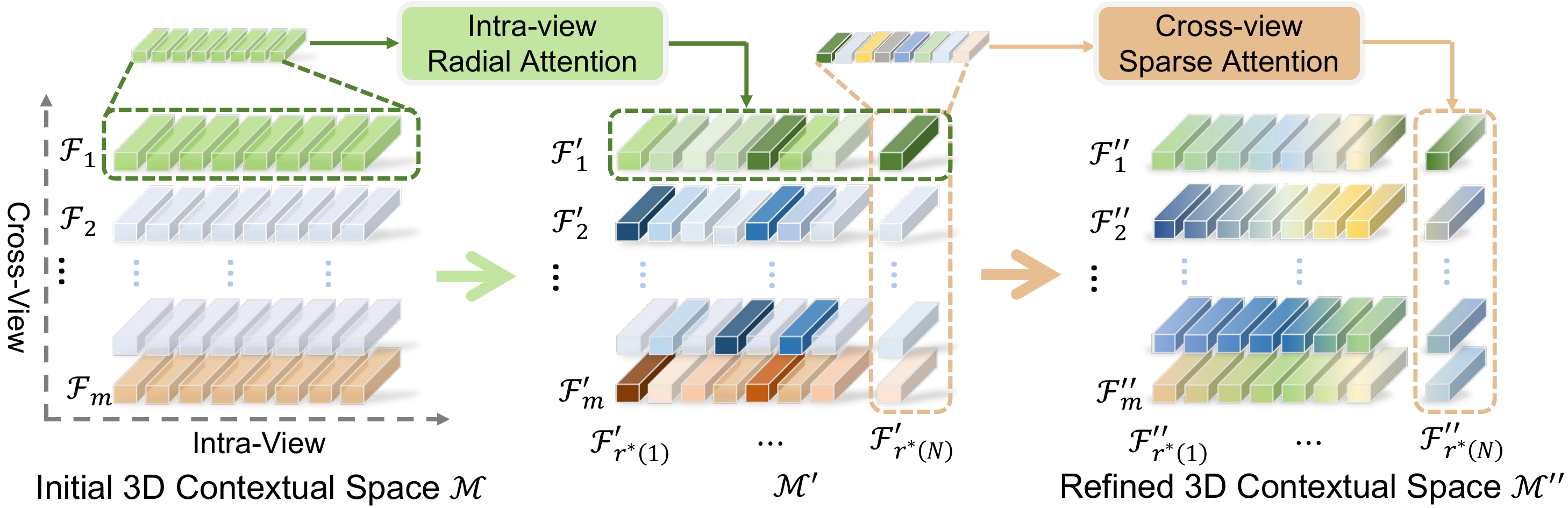}
    \vspace{-6mm}
    \caption{\textbf{Pipeline of Cross-Reprojection Attention}. Given the initial 3D contextual space $\mathcal{M}$ form Sec.~\ref{sec:contextualSpace}, we first decompose $\mathcal{M}$ along the radial direction (\ie, each intra-view). Then, we apply the \textit{intra-view radial attention module} to each $\mathcal{F}_i (i=1,...m)$ to learn the ray-aligned contextual feature from each source view and build the $\mathcal{M}^{'}$. We further decompose the $\mathcal{M}^{'}$ cross multiple views and employ the \textit{cross-view sparse attention module} to each $\mathcal{F}^{'}_{\mathbf{r}^{*}(i)}$, thus capturing sparse contextual patterns with their respective significance to semantics. After the two consecutive attention modules, we fuse the decomposed contextual information with the final refined 3D contextual space $\mathcal{M}^{''}$, which models dense semantic collections around the ray. (\textit{Best viewed in color})}
    \label{fig:cross}
\end{figure*}

\subsection{Cross-Reprojection Attention}
\label{sec:cross-reprojection}
% \noindent \textbf{Decomposing Semantic Correspondence}
% \noindent \textbf{Reassembling Semantic Connections}
To model full semantic-aware dependencies from above 3D contextual space $\mathcal{M}$, a straightforward approach is to perform dense attention over $\mathcal{M}$. However, it would suffer from heavy computational costs. To address the problem, we propose Cross-Reprojection Attention as shown in Figure~\ref{fig:cross}, including intra-view radial attention and cross-view sparse attention to approximate dense semantic connections in 3D space with lightweight computation and memory. 

\noindent \textbf{Intra-view Radial Attention.}
First, we rewrite the contextual space as $\mathcal{M} = [\mathcal{F}_1, \mathcal{F}_2, ..., \mathcal{F}_m]^T$, where $\mathcal{F}_i = [\mathcal{F}_i(\mathbf{r}^{*}(z_1)), \mathcal{F}_i(\mathbf{r}^{*}(z_2)), ..., \mathcal{F}_i(\mathbf{r}^{*}(z_N))]$ is the radial feature in the $i$-th view. 
Then, we decompose the 3D contextual space $\mathcal{M}$ along the radial direction in source views, \ie, consider the $\mathcal{F}_i (i=1,...,m)$ which encodes the intra-view contextual information within each view. Taking the $\mathcal{F}_i \in \mathbb{R}^{N \times C}$ as input, our intra-view radial attention with $H$ heads is formulated as 
\begin{equation}
    Q_R = \mathcal{F}_i W_q, \quad K_R = \mathcal{F}_i W_k, \quad V_R = \mathcal{F}_i W_v,
\end{equation}
\begin{equation}
    A^{(h)} = \sigma\big(\frac{Q_R^{(h)}K_R^{(h)T}}{\sqrt{d_k}}\big) V_R^{(h)} , h = 1,..., H  ,
\end{equation}
\begin{equation}
    f(Q_R, K_R, V_R) = \text{Concat}(A^{(1)}, ..., A^{(H)}) W_o,
\end{equation}
where $\sigma(\cdot)$ denotes the softmax function, and $d_k = C/H$ is the dimension of each head. $A^{(h)}$ denotes the embedding output from the $h$-th attention head, $Q_R^{(h)}, K_R^{(h)}, V_R^{(h)} \in \mathbb{R}^{N \times d_k}$ denote query, key, and value of radial embeddings respectively. $W_q, W_k, W_v, W_o \in \mathbb{R}^{C\times C}$ are the projection matrices. Then, we obtain a refined $\mathcal{F}^{'}_i$ as 
\begin{equation}
    \mathcal{F}^{'}_i = f(Q_R, K_R, V_R),
\end{equation}
which contains global semantic-aware patterns along the reprojected ray in $i$-th view. Similarly, we apply this intra-view radial attention module to each $\mathcal{F}_i (i=1,...,m)$ to refine the 3D contextual space, denoted as $\mathcal{M}^{'} = [\mathcal{F}^{'}_1, \mathcal{F}^{'}_2, ..., \mathcal{F}^{'}_m]^T$.

\noindent \textbf{Cross-view Sparse Attention.}
After the intra-view radial attention module, we decompose $\mathcal{M}^{'}$ cross multiple views and rewrite $\mathcal{M}^{'} = [\mathcal{F}^{'}_{\mathbf{r}^{*}(1)}, ..., \mathcal{F}^{'}_{\mathbf{r}^{*}(N)}]$, where $\mathcal{F}^{'}_{\mathbf{{r}^{*}(i)}} = [\mathcal{F}_1^{'}(\mathbf{r}^{*}(z_i)),...,\mathcal{F}_m^{'}(\mathbf{r}^{*}(z_i))]^T$, which encodes the global ray-based feature in each view. Aiming to exploit semantic information from multiple views with their respective significance which is sparse, we put $\mathcal{M}^{'}$ to the cross-view sparse attention module. Following the predefined notation, we compute the $\mathcal{F}^{''}_{\mathbf{{r}^{*}(i)}}$ from
\begin{equation}
    \mathcal{F}^{''}_{\mathbf{{r}^{*}(i)}} = f(\mathcal{F}^{'}_{\mathbf{{r}^{*}(i)}} \widetilde{W_q} , \mathcal{F}^{'}_{\mathbf{{r}^{*}(i)}} \widetilde{W_k} , \mathcal{F}^{'}_{\mathbf{{r}^{*}(i)}} \widetilde{W_v} ),
\end{equation}
where $\widetilde{W_q}, \widetilde{W_k}, \widetilde{W_v}$ are the projection matrices in cross-view sparse attention. Therefore, we get our final 3D contextual space $\mathcal{M}^{''} = [\mathcal{F}^{''}_{\mathbf{r}^{*}(1)}, ..., \mathcal{F}^{''}_{\mathbf{r}^{*}(N)}]$, which collects dense semantic connections around the query ray. 

%\noindent \textbf{discussion}

\subsection{Semantic Ray}
\label{sec:semantic-ray}
\noindent \textbf{Semantic Ray Construction.} As done in the previous pipeline, we have built a semantic-aware space $\mathcal{M}^{''} = [\mathcal{F}^{''}_1, \mathcal{F}^{''}_2, ..., \mathcal{F}^{''}_m]^T$ which encodes refined 3D contextual patterns around the light ray emitted from the query view. To construct our final semantic ray from $\mathcal{M}^{''}$ and better learn the semantic consistency along the ray, we introduce a \textit{Semantic-aware Weight Network} to rescore the significance of each source view. Then, we can assign distinct weights to different views and compute the final semantic ray $\mathbf{s}$ as
\begin{equation}
    \mathbf{s} = w_1 \mathcal{F}_1^{''} + w_2 \mathcal{F}_2^{''} + ... + w_m \mathcal{F}_m^{''},
\end{equation}
\small
\begin{equation}
     \mathbf{w} \in \mathbb{C}(\tau):=\left\{\mathbf{w}: 0<\frac{\tau}{m}<w_i<\frac{1}{\tau m}, \sum_{i=1}^m w_i=1\right\},
\end{equation}
\normalsize
where $\mathbf{w}$ is the view reweighting vector with length $m$ indicating the importance of source views, and $\tau$ is the small constant with $\tau > 0$. The deviation of the weight distribution from the uniform distribution is bound by the hyperparameter $\tau$, which keeps the semantic consistency instead of bias across views.

\noindent \textbf{Semantic Field Rendering.} Finally, we use the rendering scheme introduced in NeRF to render semantic logits from the ray $\mathbf{r}$ with N sampled points, namely $\{ z_k\}_{k=1}^N$. The semantic logit $\hat{\mathbf{S}}(\mathbf{r})$ is defined as
\begin{equation}
    \hat{\mathbf{S}}(\mathbf{r}) = \sum_{k=1}^{N} T(z_k) \{1- \text{exp}(-\sigma(z_k) \delta_k)\} \mathbf{s}(z_k),
\end{equation}
where $\quad T(z_k) = \exp(- \sum_{k^{'}=1}^{k-1} \sigma(z_k) \delta_k)$, $\delta_k = z_{k+1} - z_{k}$ is the distance between two adjacent quadrature points along the semantic ray and $\sigma$ is predicted by a \textit{Geometry-aware Network}.

\begin{table*}[!t]

	\centering
        \resizebox{\textwidth}{!}{
	\begin{tabular}{lccccccc}
		\toprule
		\multirow{2}{*}{Method} & \multirow{2}{*}{Settings} & \multicolumn{3}{c}{Synthetic Data (Replica \cite{replica}) } & \multicolumn{3}{c}{Real Data (ScanNet \cite{scannet})  } \\
		
		\cmidrule(lr){3-5}\cmidrule(lr){6-8}
		
		&& \multicolumn{1}{c}{mIoU$\uparrow$} &\multicolumn{1}{c}{Total Acc$\uparrow$} &\multicolumn{1}{c}{Avg Acc$\uparrow$} &\multicolumn{1}{c}{mIoU$\uparrow$} &\multicolumn{1}{c}{Total Acc$\uparrow$} &\multicolumn{1}{c}{Avg Acc$\uparrow$}\\

        \midrule
            MVSNeRF~\cite{mvsNeRF} + Semantic Head  & \multirow{3}{*}{\shortstack{Generalization}}  & 23.41 & 54.25 & 33.70 & 39.82 & 60.01 & 46.01 \\
		  NeuRay~\cite{neuray} + Semantic Head  &  & 35.90 & 69.35 & 43.97 & 51.03 & 77.61 & 57.12 \\
		  \textbf{S-Ray} (\textbf{Ours}) & & \textbf{41.59} & \textbf{70.51} & \textbf{47.19} & \textbf{57.15} & \textbf{78.24} & \textbf{62.55} \\

		\midrule
		 Semantic-NeRF~\cite{semantic-nerf}    &  \multirow{4}{*}{\shortstack{Finetuning}} & 75.06 & 94.36 & 70.20 & \textbf{91.24} & 97.54 & 93.89  \\ % 5000 steps
		MVSNeRF \cite{mvsNeRF} + Semantic Head$_{ft}$ &  & 53.77  & 79.48 & 62.85 &  55.26 & 76.25 & 69.70 \\

        NeuRay\cite{neuray} + Semantic Head$_{ft}$ &  & 63.73  & 85.54 & 70.05 & 77.48 & 91.56 & 81.04 \\
		\textbf{S-Ray}$_{ft}$ (\textbf{Ours})  &  & \textbf{75.96} & \textbf{96.38} & \textbf{80.81} & 91.08 & \textbf{98.20} & \textbf{93.97} \\  % 5000 steps

		\bottomrule
	\end{tabular}}
\rule{0pt}{0.01pt}
\vspace{-3mm}
\caption{\textbf{Quantitative comparison.} We show averaged results of mIoU, Total Acc, and Average Acc (higher is better) as explained in Sec.~\ref{sec:exp_set}. On the top, we compare S-Ray (Ours) with NeuRay~\cite{neuray}+semantic head and MVSNeRF~\cite{mvsNeRF}+semantic head with direct network inference. On the bottom, we show our results with only 10 minutes of optimization. }
\label{tb:rendering}
\end{table*}

\noindent \textbf{Network Training.} More specifically, we discuss the formulation of the semantic loss functions. We apply our S-Ray with a set of RGB images with known camera parameters denoted by $\mathcal{I}$. 
The losses are computed for the set of all rays denoted as $\mathcal{R}$, which is emitted from the query image $I \in \mathcal{I}$. The semantic loss is computed as multi-class cross-entropy loss to encourage the rendered semantic labels to be consistent with the provided labels, where $1 \leq l \leq L$ denotes the class index
\begin{equation}
    \mathcal{L}_{sem}(I)=-\sum_{\mathbf{r} \in \mathcal{R}}\left[\sum_{l=1}^L p^l(\mathbf{r}) \log \hat{p}^l(\mathbf{r})\right],
    \label{eq:Lsem}
\end{equation}
where $p^{l}, \hat{p}^l$ are the multi-class semantic probability at class $l$ of the ground truth map. Unlike Semantic-NeRF \cite{semantic-nerf} which needs heavy prior training for the radiance only in a single scene, we can train our S-Ray Network with semantic supervision in multiple scenes simultaneously and fast generalizes to a novel scene. 

\subsection{Discussion and Implementation}
\noindent \textbf{Discussion with GPNR~\cite{GPNR}. }
The recent work GPNR~\cite{GPNR} shares a similar motivation by aggregating features along epipolar line. It is proposed for color rendering with carefully-designed positional encoding to encode information of view direction, camera pose, and location. In contrast, we focus on learning a generalizable semantic field for semantic rendering through merely image features. In this sense, S-Ray creates a neat design space without any positional encoding engineering. While GPNR requires training 24 hours on 32 TPUs, S-Ray only needs a single RTX3090-Ti GPU with similar training time. 

\noindent \textbf{Implementation details. }
Given multiple views of a scene, we construct a training pair of source and query view by first randomly selecting a target view, and sampling $m$ nearby but sparse views as source views. We follow \cite{neuray} to build our sampling strategy, which simulates various view densities during training, thus helping the network generalize across view densities. We implement our model in PyTorch \cite{pytorch} and train it end-to-end on a single RTX3090-Ti GPU with 24GB memory. The batch size of rays is set to 1024 and our S-Ray is trained for $250k$ steps using Adam \cite{Adam} with an initial learning rate of $1e-3$ decaying to $5e-5$. S-Ray is able to generalize well to novel scenes and can also be finetuned per scene using the same objective in (\ref{eq:Lsem}). More details of network training, architecture design, and hyperparameter settings can be found in the supplementary.

\section{Experiments}
We conduct extensive experiments and evaluate our method with basically two settings. 1) We directly evaluate our pretrained model on test scenes (\ie, unseen scenes) without any finetuning. Note that we train only \textit{one} model called S-Ray and evaluate it on all test scenes. 2) We finetune our pretrained model for a small number of steps on each unseen scene before evaluation. While training from scratch usually requires a long optimization time, we evaluate our S-Ray by achieving comparable performance with well-trained models by much less finetuning time.
\subsection{Experiment Setup}
\label{sec:exp_set}
\noindent \textbf{Datasets.} To test the effectiveness of our method comprehensively, we conduct experiments on both synthetic data and real data. For synthetic data, we use the Replica \cite{replica}, a dataset with 18 highly photo-realistic 3D indoor scene reconstructions at room and building scale. Each Scene consists of dense geometry, high-dynamic-range textures, and per-primitive semantic class. Then, we choose 12 scenes from the Replica as training datasets and the remaining unseen scenes as test datasets. For Real data, we use the ScanNet \cite{scannet}, which is a real-world large labeled RGB-D dataset containing 2.5M views in 1513 scenes annotated with 3D camera poses surface reconstructions and semantic segmentation. We choose 60 different scenes as training datasets and 10 unseen novel scenes as test datasets to evaluate generalizability in real data. More details about the split of datasets will be shown in the supplementary.

\noindent \textbf{Metrics.} To accurately measure the performance of our S-Ray, we adopt mean Intersection-over-Union (mIoU) as well as average accuracy and total accuracy to compute segmentation quality. What's more, we will discuss our rendering quality if we compute the color from the \textit{Geometry-Aware Network} in the posterior subsection. To evaluate rendering quality, we follow NeRF~\cite{NeRF}, adopting peak signal-to-noise ratio (PSNR), the structural similarity index measure (SSIM)~\cite{ssim}, and learned perceptual image patch similarity (LPIPS) \cite{lpips}. 

\begin{figure*}[!t]
    \centering
    \includegraphics[width=\linewidth]{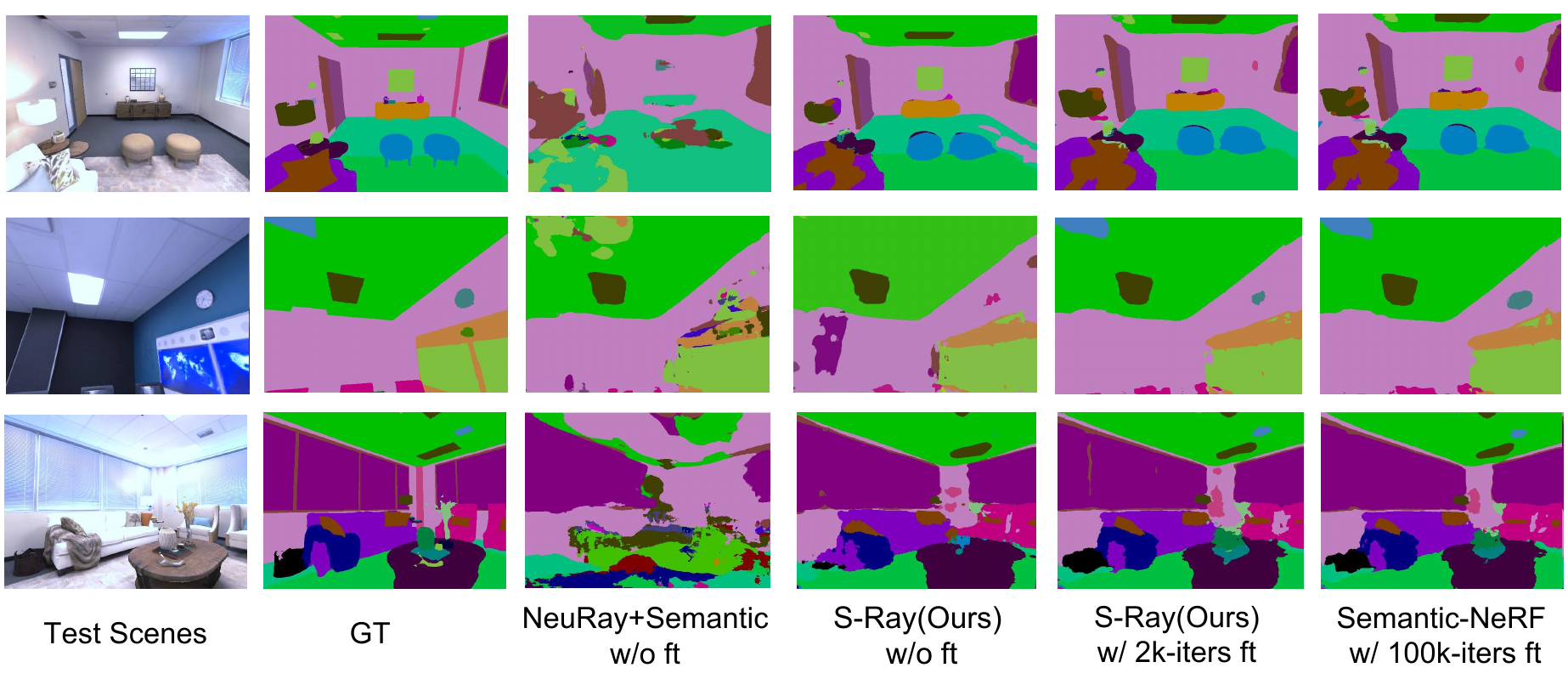}
    \vspace{-8mm}
    \caption{\textbf{Semantic rendering quality comparison}. On the left, we show direct semantic rendering results of our method and NeuRay~\cite{neuray}+semantic head. Limited by insufficient generalization, NeuRay+semantic head has difficulty to render semantics in unseen scenes and fails to capture contextual structure, while our method is able to learn the semantic structural prior, thus showing strong generalizability across different scenes. On the right, we show the experimental comparison between our S-Ray with $2k$ iterations finetuning ($\sim$10min) and Semantic-NeRF \cite{semantic-nerf} with $100k$ iterations. }
    \label{fig:result}
\end{figure*}

\noindent \textbf{Baselines.} To evaluate our fast generalizability in an unseen scene, we choose Semantic-NeRF \cite{semantic-nerf} as our baseline. Since we are the first to learn a generalizable semantic field in real-world scenes, we have no baselines to compare our generalizable performance. In order to further show our strong generalizability, we add the semantic head by following Semantic-NeRF settings to the NeRF-based methods which have shown generalizability in the reconstruction task. Specifically, we compare our method against MVSNeRF \cite{mvsNeRF} with semantic head and NeuRay \cite{neuray} with semantic head in generalization and finetuning. Due to the space limitation, We provide a more detailed discussion and comparison with~\cite{mvsNeRF, neuray, IBRnet, semantic-nerf, GPNR} in the supplementary.

\subsection{Comparison with Baselines}
To render each semantic map of the test view, we sample 8 source views from the training set for all evaluation datasets in generalization settings. For the per-scene optimization, we follow \cite{semantic-nerf} to train it on the new scene from scratch. To compare our results fairly, we follow the Semantic-NeRF \cite{semantic-nerf} to resize the images to $640\times 480$ for Replica \cite{replica} and $320\times 240$ for ScanNet \cite{scannet}. Results can be seen in Table~\ref{tb:rendering} and in Figure~\ref{fig:result}.

Table~\ref{tb:rendering} shows that our pretrained model generalizes well to unseen scenes with novel contextual information. we observe that the generalization ability of our S-Ray consistently outperforms both NeuRay \cite{neuray} and MVSNeRF \cite{mvsNeRF} with semantic heads. Although they are the recent methods that have strong generalization ability, we show that directly following Semantic-NeRF by adding a semantic head fails to fully capture the semantic information. Instead, our Cross-Reprojection Attention can extract relational features from multi-view reprojections of the query ray, thus achieving better accuracy and stronger generalization ability. 

After finetuning for only 10 minutes, our performance is competitive and even better than Semantic-NeRF with $100k$ iters per-scene optimization. The visual results in Figure~\ref{fig:result} clearly reflect the quantitative results of Table~\ref{tb:rendering}. The generalization ability of our S-Ray is obviously stronger than NeuRay \cite{neuray} with semantic head. As MVSNeRF \cite{mvsNeRF} shows even worse generalization performance than NeuRay with the semantic head as shown in Table~\ref{tb:rendering}, we do not show its visualization results due to the limited space. In general, the comparison methods directly use point-based features which benefit to per-pixel reconstruction instead of semantics. Our approach utilizes the full spatial context information around the query ray to build our semantic ray, thus leading to the best generalization ability and high segmentation across different test scenes.

\subsection{Ablation Studies and Analysis}
\begin{figure}[tb]
    \centering
    \includegraphics[width=0.98\linewidth]{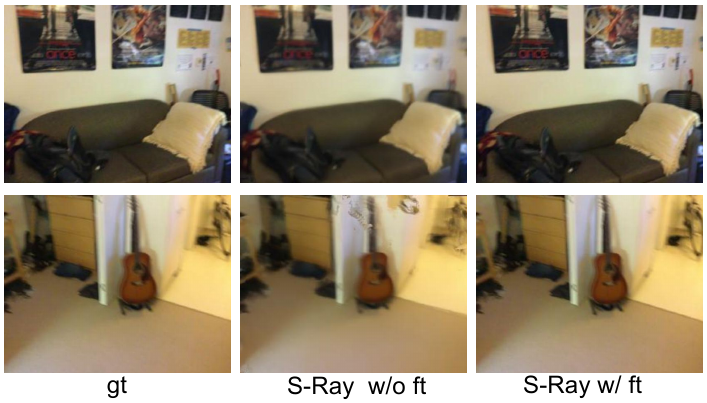}
    \vspace{-3mm}
    \caption{Qualitative results of scene rendering for generalization (w/o ft) and finetuning (w/ ft) settings in real data~\cite{scannet}.}
    \label{fig:color_result}
\end{figure}

\noindent \textbf{Reconstruction quality of S-Ray.} To evaluate the reconstruction quality of S-Ray, we follow (\ref{eq:render_discrete}) to render the radiance from geometry aware network with photometric loss same as \cite{semantic-nerf}. In Table~\ref{tab:render_quality}, we test the S-Ray for color rendering and compare it with Semantic-NeRF\cite{semantic-nerf}, MVSNeRF \cite{mvsNeRF} and NeuRay~\cite{neuray}. More comparisons can be found in the supplementary. As shown in Table \ref{tab:render_quality} and Figure \ref{fig:color_result}, our S-Ray can also generalize well in reconstruction. This shows that our network cannot only capture the contextual semantics but also learn geometry features well.

\noindent \textbf{Training from scratch.} In order to further present the generalization ability of our S-Ray, we train our model on a scene from scratch without pretraining the model. We strictly follow the same process as Semantic-NeRF \cite{semantic-nerf} to train S-Ray. Results in Table \ref{tab:ablation} (ID 9 and 10) show that training our method from scratch can also achieve similar results as finetuning the pretrained model, and it obtains even better results than Semantic-NeRF.
\begin{table}[tb]
    \centering
    \small
%\begin{footnotesize}
    \begin{tabular}{l |  c  c  c }
% \hline
\toprule 
Method & PSNR$\uparrow$ & SSIM$\uparrow$ & LIPIPS$\downarrow$ \\  % ScanNet Scene0101_03
\midrule
Semantic-NeRF~\cite{semantic-nerf}  & 25.07 & 0.797 & 0.196\\

MVSNeRF~\cite{mvsNeRF}  & 23.84 & 0.733 & 0.267\\

NeuRay~\cite{neuray}  & 27.22 & 0.840 & 0.138\\\

\textbf{S-Ray} (\textbf{Ours})  & 26.57 & 0.832 & 0.173 \\

\textbf{S-Ray}$_{ft}$ (\textbf{Ours})  & \textbf{29.27} & \textbf{0.865} & \textbf{0.127}\\

\bottomrule  
\end{tabular}
%\end{footnotesize}
\vspace{-3mm}
\caption{Comparisons of scene rendering in real data~\cite{scannet}.}
     \label{tab:render_quality}
\end{table} 

\noindent \textbf{Evaluation of Cross-Reprojection Attention module.} In Table \ref{tab:ablation}, we adopt ablation studies for Cross-Reprojection module shown in Figure \ref{fig:cross}. We compare four models, the full model (ID 1), the model without Cross-Reprojection Attention (ID 2, 4), the model only with intra-view radial attention module (ID 3, 7), and the model only with cross-view attention (ID 4, 8). The results show that Cross-Reprojection Attention in S-Ray enables our network to learn more contextual information as discussed in Sec.~\ref{sec:cross-reprojection}.

\begin{table}[!h]
    \centering
    \resizebox{\linewidth}{!}{
    \begin{tabular}{ccccc}
        \toprule
        ID & Description & Setting & mIoU & Total Acc \\
        \midrule
        1 & full S-Ray                                & Gen & \textbf{57.15}  & \textbf{78.24}  \\
        2 & w/o cross-reprojection Att             & Gen &  45.34 & 53.67 \\
        3 & only intra-view Att & Gen & 49.09 & 58.53 \\
        % 4 & $M,B,P$ with init-NeuRay-D & Gen &  29.79 & 24.69 \\
        4 & only cross-view Att    & Gen & 52.56 & 63.25 \\
        \midrule
        5 & full S-Ray      & Ft & \textbf{91.08} & \textbf{98.20} \\
        6 & w/o cross-reprojection Att & Ft & 76.30 & 86.02 \\
        7 & only intra-view Att  & Ft & 81.24 & 89.58 \\
        8 &  only cross-view Att  & Ft & 87.01 & 93.34 \\
        \midrule
        9 &  S-Ray  & Sc & \textbf{95.31}  &  \textbf{98.40} \\
        10 &  Semantic-NeRF~\cite{semantic-nerf} & Sc & 94.48 & 95.32\\
        \bottomrule
    \end{tabular}
    }
    \vspace{-3mm}
    \caption{Ablation studies. mIoU and Total Acc on the real data from ScanNet~\cite{scannet}. ``Gen'' means the generalization setting, ``Ft'' means to finetune on the scene and ``Sc'' means to train from scratch.}
    \label{tab:ablation}
\end{table}

\noindent \textbf{Few-step finetuning of S-Ray.} Table \ref{tab:time_compare} reports mIoU and finetuning time of different models with on the ScanNet \cite{scannet} dataset. We observe that by finetuning with limited time, our model is able to achieve a better performance than a well-trained Semantic-NeRF \cite{semantic-nerf} with much longer training time. As Semantic-NeRF fails to present cross-scene generalization ability, it still requires full training on the unseen scene. Instead, our S-Ray is able to quickly transfer to the unseen scenes. Moreover, S-Ray outperforms other competitive baselines with similar finetuning time, which further demonstrates that our Cross-Reprojection Attention operation successfully improves the generalization ability.
\begin{table}[!h]
    \centering
    \resizebox{\linewidth}{!}{
    \begin{tabular}{ccccc}
        \toprule
         Method & Train Step & Train Time & mIoU  \\
         \midrule
        %  NeRF~\cite{mildenhall2020nerf} & 10k  & $\sim$30min &       \\
         Semantic-NeRF~\cite{semantic-nerf}    & 50k & $\sim$2h  & 89.33 & \\
         MVSNeRF~\cite{mvsNeRF} w/ s-Ft & 5k  & $\sim$20min & 52.02 & \\
         NeuRay~\cite{neuray} w/ s-Ft   & 5k   & $\sim$32min & 79.23 & \\
         \textbf{S-Ray}-Ft (\textbf{Ours})                         & 5k   & $\sim$20min & \textbf{92.04} & \\
         \bottomrule
    \end{tabular}
    }
    \vspace{-3mm}
    \caption{mIoU and training steps/time on real data~\cite{replica}. ``w/ s'' means adding a semantic head on the baseline architectures. }
    \label{tab:time_compare}
\end{table}

From the experiments above, we have the following key observations:
\begin{enumerate}[1)]
    \item Our Semantic Ray can exploit contextual information of scenes and  presents strong generalization ability to adapt to unseen scenes. It achieves encouraging performance without finetuning on the unseen scenes, and also obtains comparable results to the well-trained Semantic-NeRF with much less time of finetuning.
    \item We show the effectiveness of our Cross-Reprojection Attention module through comprehensive ablation studies. Experiments demonstrate that both intra-view and cross-view attentions are crucial for S-Ray, and we achieve the best performance by simultaneously exploiting relational information from both modules.
    \item Our performance in radiance reconstruction shows the great potential of our attention strategy, which is able to learn both dense contextual connections and geometry features with low computational costs. 
\end{enumerate}
\section{Conclusion}
In this paper, we have proposed a generalizable semantic field named Semantic Ray, which is able to learn from multiple scenes and generalize to unseen scenes. Different from Semantic NeRF which relies on positional encoding thereby limited to the specific single scene, we design a Cross-Reprojection Attention module to fully exploit semantic information from multiple reprojections of the ray. In order to capture dense connections of reprojected rays in an efficient manner, we decompose the problem into consecutive intra-view radial and cross-view sparse attentions to extract informative semantics with small computational costs. 
Extensive experiments on both synthetic and real-world datasets demonstrate the strong generalizability of our S-Ray and effectiveness of our Cross-Reprojection Attention module. With the generalizable semantic field, we believe that S-Ray will encourage more explorations of potential NeRF-based high-level vision problems in the future.

\noindent \textbf{Acknowledgements. }This work was supported in part by the National Natural Science Foundation of China under Grant 62206147, and in part by Deng Feng Fund.

%%%%%%%%% REFERENCES
{\small
\bibliographystyle{ieee_fullname}
\bibliography{cvpr2023_conference}
}
\section*{Appendix}
\section*{Additional implementation details}
\noindent \textbf{Training details.} 
At the training time, we first project the query ray instead of a single point to each source view and fetch the corresponding ray-based feature, which contains rich contextual information in each intra-view. 
For pre-training, we train on a single NVIDIA RTX3090-Ti GPU with 24GB memory. On this hardware, we train our S-Ray for 260k iterations in 60 different scenes of ScanNet~\cite{scannet} (real-world data) and 100k iterations in 12 different scenes of Replica~\cite{replica}  (synthetic data). For finetuning, we only require 10min finetuning time corresponding to 2k iterations. This finetuning result is comparable and even better than 100k optimizations of Semantic-NeRF \cite{semantic-nerf} from each independent scene.

We do not show the specific details of the semantic loss design in the paper. In code implementation, we apply two-stage (coarse and fine) ray sampling as done in NeRF~\cite{NeRF}. Therefore, our semantic loss is actually computed as
\begin{equation}
    L_{sem}=-\sum_{\mathbf{r} \in \mathcal{R}}\left[\sum_{l=1}^L p^l(\mathbf{r}) \log \hat{p}_c^l(\mathbf{r})+\sum_{l=1}^L p^l(\mathbf{r}) \log \hat{p}_f^l(\mathbf{r})\right]
\end{equation}

where $\mathcal{R}$ are the set of sample rays within a training batch, $1\leq l\leq L$ is the class index, and $p^l, \hat{p}^l_c, \hat{p}^l_f$ are the multi-class probability at class $l$ of the ground truth, coarse semantic logits and fine semantic logits for the query ray $\mathbf{r}$. Actually, for fair comparison in Section 4.2 of our paper, we adopt the same training loss with Semantic-NeRF~\cite{semantic-nerf} as:
\begin{equation}
    \mathcal{L}_{total} = \lambda_1 \mathcal{L}_{sem} + \lambda_2 \mathcal{L}_{photometric}, 
\end{equation}
where the color head is from the geometry aware network with photometric loss same as~\cite{semantic-nerf}. Like Semantic-NeRF, we also set $\lambda_1 = \lambda_2 =1$ in Section 4.2 and set $\lambda_1 = 0, \lambda_2 = 1$ as NeRF for ablation study in Table 2 of the paper. 

\noindent \textbf{Data split.} Our training data consists of both synthetic data and real data. For real data training, we choose 60 different scenes from ScanNet~\cite{scannet} as training datasets and use the image resolution of $320 \times 240$. We then choose 10 unseen novel scenes as test datasets to evaluate the generalizability of S-Ray in real data. For synthetic data, we choose 12 different scenes (\ie, 2 rooms, 2 offices, 7 apartments, 1 hotel) from Replica~\cite{replica} for the training set and the remains (\ie, 2 apartments, 3 offices, 1 room) as test set with the image resolution of $640\times 480$. 
For each test scene, we select 20 nearby views; we then select 8 views as source input views, 8 as additional input for per-scene fine-tuning, and take the remaining 4 as testing views. Our training data includes various camera setups and scene types, which allows our method to generalize well to unseen semantic scenarios.

\section*{Additional experiments and analysis}
\begin{figure}[!h]
    \centering
    \includegraphics[width=\linewidth]{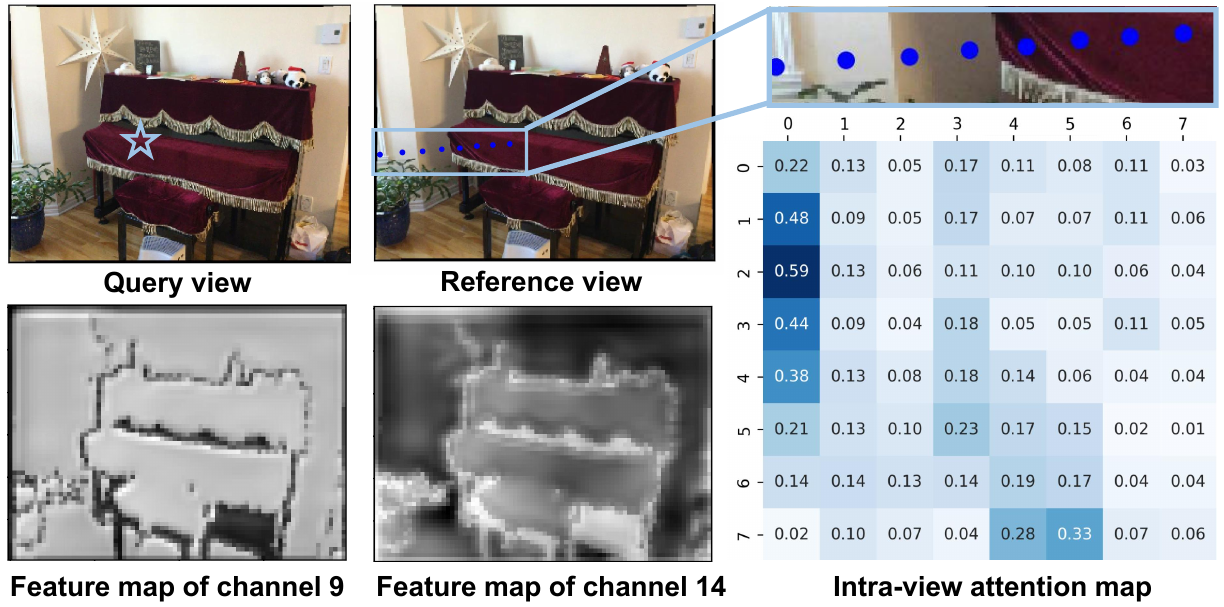}
    \vspace{-6mm}
    \caption{Visualization of 2D CNN features from ResUnet and intra-view attention map. It shows that our ResUnet can help S-Ray learn reasonable geometry for contextual semantics and the intra-view attention map is closely related to the visibility.}
    \label{fig: vis}
\end{figure}

\noindent \textbf{More discussion of loss function.} When adding color rendering, it is interesting to see the effect of the weighting factor, thus conducting the following experiments in Table~\ref{tab:weight-factor}. We observe that color rendering can benefit semantics but color rendering is not sensitive to semantics. Furthermore, Table~\ref{tab:weight-factor} shows that the semantic loss alone can also drive our model to learn reasonable contextual geometry for semantic information as visualized in Figure~\ref{fig: vis}.

\begin{table}[!h]
\centering
\resizebox{\columnwidth}{!}{%
\begin{tabular}{cccccc}
\hline
$\lambda_1/\lambda_2$ & 1/0   & 0.75/0.25 & 0.5/0.5 & 0.25/0.75 & 0/1   \\ \hline
PSNR                  & 17.49 & 25.26     & 25.35   & 26.24     & 26.57 \\
mIoU(\%)              & 55.10 & 56.51     & 57.15   & 58.12     & 3.62  \\ \hline
\end{tabular}%
}
\vspace{-3mm}
\caption{Different weighting factors effect under ScanNet~\cite{scannet} generalization settings. }
\label{tab:weight-factor}
\end{table}

\noindent \textbf{Effectiveness of the CRA module.} To further validate the computational effectiveness of our Cross-Reprojection Attention (CRA) module, we provide the comparisons with Dense Attention in FLOPs and Memory usage.

\begin{table}[!h]
    \centering
    \resizebox{\linewidth}{!}{
    \begin{tabular}{lcccc}
        \toprule
         Description & GFLOPs & mIoU(\%) &  Total Acc(\%)\\
         \midrule
        %  NeRF~\cite{mildenhall2020nerf} & 10k  & $\sim$30min &       \\
        w/o CRA      & 0 & 76.30  & 86.02 & \\
         Dense Attention &  10.25 & 90.46  & 94.52  & \\
         only intra-view Att  & 3.05  & 81.24 & 89.58 & \\
         only cross-view Att  & 2.35  & 87.01 & 93.34 & \\
         full CRA     & \textbf{5.40} & \textbf{91.08}  & \textbf{98.20} & \\
         \bottomrule
    \end{tabular}
    }
    \vspace{-3mm}
    \caption{Performance on real data \cite{scannet} for different settings of Cross-Reprojection Attention module (CRA). FLOPs increments are estimated for the input of $1024 \times 64 \times 8 \times 32$. }
    \label{tab:flop_compare}
\end{table}

\begin{table}[!h]
    \centering
    \resizebox{\linewidth}{!}{
    \begin{tabular}{lcccc}
        \toprule
         Description & Memory(MB) & mIoU(\%) &  Total Acc(\%)\\
         \midrule
        %  NeRF~\cite{mildenhall2020nerf} & 10k  & $\sim$30min &       \\
        w/o CRA      & 0 &  76.30 &86.02  & \\
         Dense Attention &  17647 & 90.46 & 94.52 & \\
         only intra-view Att  & 3899  & 81.24 & 89.58 & \\
         only cross-view Att  &  1583 & 87.01 & 93.34 & \\
          full CRA     & \textbf{4143} & \textbf{91.08}  & \textbf{98.20} & \\
         \bottomrule
    \end{tabular}
    }
    \vspace{-3mm}
    \caption{Performance on real data \cite{scannet} for different settings of Cross-Reprojection Attention module (CRA). Memory increments are estimated for an input of $1024 \times 64 \times 8 \times 32$.}
    \label{tab:mem_compare}
\end{table}

Table~\ref{tab:flop_compare} and Table~\ref{tab:mem_compare} show the computational performance of real data by adopting different settings of our Cross-Reprojection Attention (CRA) module. We observe that directly applying the dense attention over multi-view reprojected rays suffers from heavy computational cost and high GPU memory. In contrast, our CRA module can achieve the comparable performance of dense attention with friendly GPU memory and high computational efficiency. Specifically, the design of CRA can improve the performance by $47.3\%$ in FlOPs and $76.5\%$ in GPU memory. These results prove that the proposed cross-reprojection attention can achieve high mIoU and total accuracy by capturing dense and global contextual information with computational efficiency.
\newline
\newline
\noindent \textbf{Semantic ray construction.} To construct the final semantic ray in Section 3.4 of our paper, we assign distinct weights to different source views and compute the semantic ray with semantic consistency. Specifically, we design the \textit{Semantic-aware Weight Network} to rescore the significance of each view with a hyperparameter $\tau$, as 
\begin{equation}
     \mathbf{w} \in \mathbb{C}(\tau):=\left\{\mathbf{w}: 0<\frac{\tau}{m}<w_i<\frac{1}{\tau m}, \sum_{i=1}^m w_i=1\right\},
\end{equation}
where $\mathbf{w}$ is the view reweighting vector with length $m$ indicating the importance of source views. Instead of mean aggregation which ignores the different significance of different source views, the hyperparameter $\tau$ controls the semantic awareness of each view. The effectiveness of $\tau$ can be seen in Table~\ref{tab:tau}.

\begin{table}[!h]
    \centering
    \resizebox{\linewidth}{!}{
    \begin{tabular}{lcccc}
        \toprule
         hyperparameter $\tau$ & mIoU(\%) & Total Accuracy(\%)  & Average Accuracy(\%)\\
         \midrule
        %  NeRF~\cite{mildenhall2020nerf} & 10k  & $\sim$30min &       \\
        1    & 54.21 &   77.13  &  59.05\\
         0.8 &  56.33 & 78.01 & 60.37& \\
         0.7  & \textbf{57.15} & \textbf{78.24} & \textbf{62.55} & \\
         0.5 & 55.70  & 76.64 & 60.80 & \\
          0.2  &54.03  & 77.25 &  61.34 & \\
         \bottomrule
    \end{tabular}
    }
    \vspace{-3mm}
    \caption{Performance on real data \cite{scannet} for different settings of hyperparameter $\tau$ in test set.}
    \label{tab:tau}
\end{table}

\begin{figure*}[!h]
    \centering
    \includegraphics[width=\linewidth]{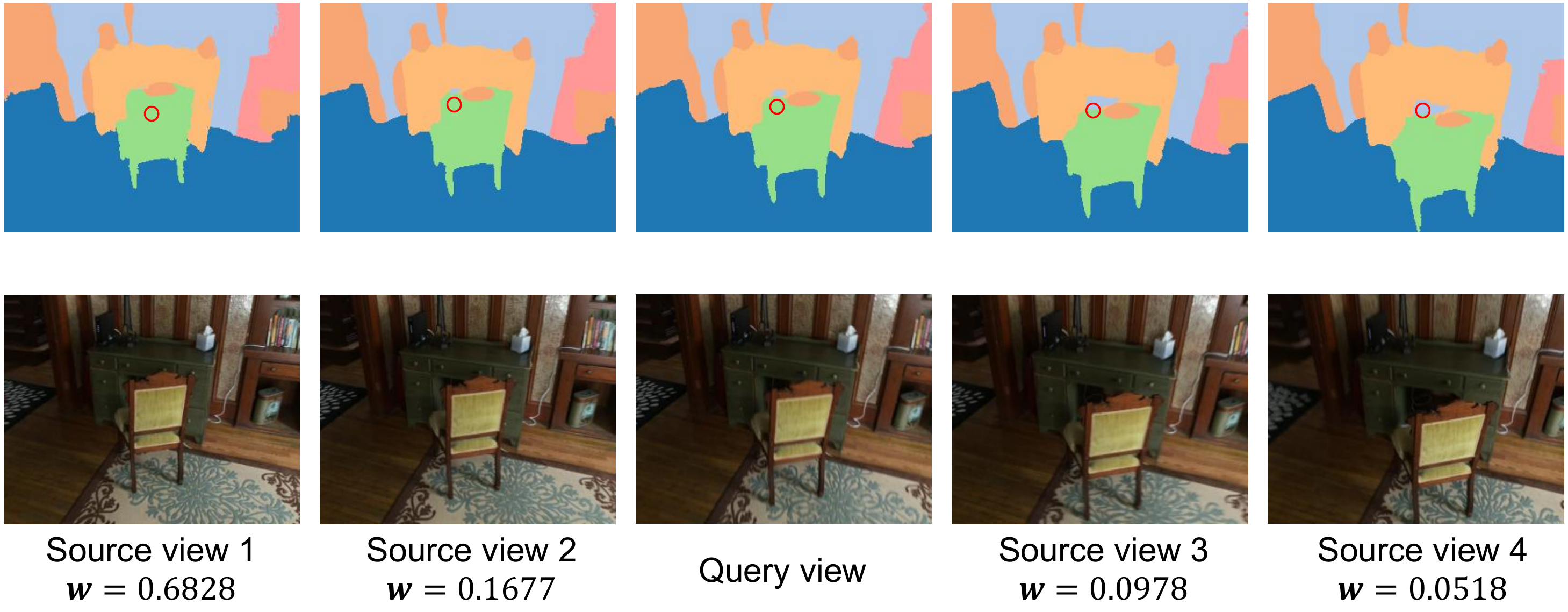}
    \caption{Different significance weight of source view. Given the query ray, we apply the semantic-aware weight network to learn the significance weight $\mathbf{w}$ to restore each source view. Note that the greater weight will be assigned to the more important source view. }
    \label{fig:new_weight}
\end{figure*}

From Table~\ref{tab:tau}, we observe that we can improve the performance of semantic segmentation by assigning different weights to each source view with hyperparameter $\tau$. Note that $\tau = 1$ means the mean aggregation operation.
\newline
\newline
\noindent \textbf{Training process.}  Given multiple views of a scene, we construct a training pair of source and query view (\ie, target view) by first randomly selecting a target view, and sampling 8 nearby but sparse views as source views. We follow \cite{neuray} to build our sampling strategy. The performance of our method in different training iterations can be found in Table~\ref{tab:training_detail}. The results show that we only require 260k iterations for 20 hours to train our S-Ray over 60 different real scenes, which demonstrates the efficiency and effectiveness of our network design.
\newline
\newline
\noindent \textbf{More comparisons with Semantic-NeRF.} To further show our strong and fast generalizability in a novel unseen scene, we compare our performance with Semantic-NeRF~\cite{semantic-nerf} in per-scene optimization. The results are shown in Table~\ref{tab:compare_semanticnerf}. While Semantic-NeRF~\cite{semantic-nerf} needs to train one independent model for an unseen scene, we observe that our network S-Ray can effectively generalize across unseen scenes. What's more, our direct result can be improved by fine-tuning on more images for only 10 min (2k iterations), which achieves comparable quality with Semantic-NeRF for 100k iterations per-scene optimization. Moreover, Semantic-NeRF shows very limited generalizability by first generating pseudo semantic labels for an unseen scene with a pretrained model, and then training on this scene with the pseudo labels. In this way, Semantic-NeRF is able to apply to new scenes without GT labels. In contrast, our S-Ray provides stronger generalization ability by enabling directly test on unseen scenes. We provide additional experiments in Table~\ref{tab:pseudo}.
\newline
\newline
\noindent \textbf{Comparison with GPNR.} The recent work GPNR~\cite{GPNR} also generates novel views from unseen scenes by enabling cross-view communication through the attention mechanism, which makes it a bit similar to our S-Ray. To further justify the motivation and novelty, we summarize several key differences as follows. \textbf{Tasks:} GPNR utilizes fully attention-based architecture for color rendering while our S-Ray focuses on learning a generalizable semantic field for semantic rendering. \textbf{Embeddings:} GPNR applies three forms of positional encoding to encode the information of location, camera pose, view direction, etc. In contrast, our proposed S-Ray only leverages image features with point coordinates without any handcrafted feature engineering. In this sense, our S-Ray enjoys a simpler design in a more efficient manner. \textbf{Training cost.} While GPNR requires training 24 hours on 32 TPUs, S-Ray only needs a single RTX3090-Ti GPU with similar training time.
\begin{table}[!h]
\centering
\resizebox{\columnwidth}{!}{%
\begin{tabular}{cccccccc}
\hline
       & w/o ft & ft 5k(p) & ft 5k(gt) & ft 50k(p) & ft 50k(gt) & ft converge(p) & ft converge(gt) \\ \hline
S-NeRF & N/A    & 78.59    & 86.32     & 85.64     & 91.33      & 92.10          & 95.24           \\
S-Ray  & 77.82  & 88.07    & 93.40     & 91.25     & 95.15      & 92.43          & 95.39           \\ \hline
\end{tabular}%
}
\vspace{-3mm}
\caption{More mIoU comparisons with SemanticNeRF(S-NeRF) in the scene0160-01 from ScanNet. Same with S-NeRF, we choose pretrained DeepLabV3+~\cite{deeplabv3} to generate pseudo semantic labels for finetuning. ``p'' means finetuning with pseudo labels, and ``gt'' means finetuning with ground truth.}

\label{tab:pseudo}
\end{table}
\newline
\newline
\noindent \textbf{More discussion for reconstruction quality.} To further demonstrate the reconstruction quality and generalizability of S-Ray, we evaluate S-Ray with NeuRay~\cite{neuray}, MVSNeRF~\cite{mvsNeRF}, and IBR-Net~\cite{IBRnet} on two typical benchmark datasets (\ie, Real Forward-facing~\cite{mildenhall2019local} and Realistic Synthetic 360$^\circ$~\cite{NeRF}) in Table~\ref{tab:benchmark}. In general, Table~\ref{tab:benchmark} shows our Cross-Reprojection Attention module is also useful for generalizable NeRFs with out semantic supervision.
\begin{table}[!h]
\centering
\resizebox{\columnwidth}{!}{%
\begin{tabular}{ccccccc}
\hline
            & \multicolumn{3}{c}{Realistic Synthetic 360°}        & \multicolumn{3}{c}{Real Forward-facing}             \\ \hline
Method      & PSNR$\uparrow$ & SSIM$\uparrow$ & LPIPS$\downarrow$ & PSNR$\uparrow$ & SSIM$\uparrow$ & LPIPS$\downarrow$ \\ \hline
MVSNeRF & 23.46 & 0.851 & 0.172 & 22.93 & 0.794 & 0.260  \\
IBRNet  & 24.52 & 0.874 & 0.158 & 24.17 & 0.802 & 0.215 \\
NeuRay  & 26.73 & 0.908 & 0.123 & 25.35 & 0.824 & 0.198 \\
S-Ray(Ours) & \textbf{26.84} & \textbf{0.917} & \textbf{0.115}    & \textbf{25.68} & \textbf{0.833} & \textbf{0.180}     \\ \hline
\end{tabular}%
}
\vspace{-3mm}
\caption{Quantitative comparisons of scene rendering in the generalization setting. All generalization methods including our method are pretrained on the same scenes and tested on unseen test scenes. }
\label{tab:benchmark}
\end{table}
While the three mentioned methods in Table~\ref{tab:benchmark} and our method are image-based rendering, the main difference lies in how to extract useful features: (a) MVS-NeRF leverages cost volume to extract geometry features, which benefits the acquisition of density; IBRNet performs feature attention on rays in 3D space and NeuRay further extracts the occlusion-aware features by explicitly modeling occlusion. Their features are sparse in 3D space but sufficient for color rendering. (b) Our method goes back to the 2D reprojection space and obtains dense attention by cascading two sparse attention modules, thus extracting rich semantic and geometric features. A key point is that we apply a ResUnet segmentation network fro context feature extraction to get semantic priors, which is not present in the previous methods.
\newline
\newline
\noindent \textbf{Disscusion of the number of source views.} 
\begin{table}[!h]
    \centering
    \resizebox{\linewidth}{!}{
    \begin{tabular}{lccccc}
        \toprule
         $N_s$ & mIoU(\%) & Total Acc(\%)  & Avg Acc(\%) & PSNR & SSIM\\
         \midrule
        %  NeRF~\cite{mildenhall2020nerf} & 10k  & $\sim$30min &       \\
        1    & 67.55 &  86.15   & 73.73 & 26.47 & 0.9077 \\
         4&  75.41 & 90.51  & 81.06 & 30.90 & 0.9368 \\
         8  & {79.97} & {93.06} & {84.92}  &  \textbf{29.52} & \textbf{0.9106}\\
         12 & {83.21} & 93.89 & 88.07 & 28.57 & 0.8969 \\
          16  & \textbf{84.84}& \textbf{94.33} & \textbf{89.78} & 27.85 & 0.8859 \\
         \bottomrule
    \end{tabular}
    }
    \vspace{-3mm}
    \caption{Performance(mIoU, Total accuracy, Average accuracy, PSNR, SSIM) on the real data scene~\cite{scannet} wiht different source view numbers $N_s$. } 
    \label{tab:Ns_views}
\end{table}
We observe that using more source views on our S-Ray model can improve semantic rendering quality. The results are shown in Table~\ref{tab:Ns_views}. The reason is that adding more reference views in training means leveraging more contextual information for semantic feature learning to build a larger 3D contextual space and reconstruct the final semantic ray, which improves the view consistency and accuracy of semantic segmentation.
\newline
\newline
\noindent \textbf{Disccusion of semantic-aware weight.} In semantic ray construction, we learn the view reweighting vector $\mathbf{w}$ to rescore the significance of each source view. To further demonstrate the effectiveness of this rescore strategy, we show the example in Figure~\ref{fig:new_weight}. The results show that $\mathbf{w}$ can distinct the different significance of different source views to the query semantic ray.

\section*{Network architecture}
\noindent \textbf{Semantic feature extraction.} Given input views and a query ray, we project the ray to each input view and apply the semantic feature extraction module in Table~\ref{table:feature_extraction} to learn contextual features and build an initial 3D contextual space. The details can be found in Table~\ref{table:feature_extraction} and Section 3.2 in the paper.
\begin{table*}[!h]
\centering

\begin{tabular}{lcccc}
\toprule
Type & Size/Channels & Activation & Stride & Normalization\\
\midrule
Input 1: RGB images & - & - & - & -\\
Input 2: View direction differences & - & - & - & -\\
L1: Conv $7\times 7$ & $3, 16$ & ReLU & $2$ & Instance\\
% layer1
L2: ResBlock $3\times 3$ & $16, 32, 32$ & ReLU & $2, 1$ & Instance\\
% layer2
L3: ResBlock $3\times 3$ & $32, 64, 64$ & ReLU & $2, 1$ & Instance\\
L4: ResBlock $3\times 3$ & $64, 64, 64$ & ReLU & $1, 1$ & Instance\\
% layer3
L5: ResBlock $3\times 3$ & $64, 128, 128$ & ReLU & $2, 1$ & Instance\\
L6: ResBlock $3\times 3$ & $128, 128, 128$ & ReLU & $1, 1$ & Instance\\
L7: ResBlock $3\times 3$ & $128, 128, 128$ & ReLU & $1, 1$ & Instance\\
L8: ResBlock $3\times 3$ & $128, 128, 128$ & ReLU & $1, 1$ & Instance\\
L9: ResBlock $3\times 3$ & $128, 128, 128$ & ReLU & $1, 1$ & Instance\\
L10: ResBlock $3\times 3$ & $128, 128, 128$ & ReLU & $1, 1$ & Instance\\

L11: Conv $3\times3$ & $128, 64$ & - & $1$ & Instance\\
L12: Up-sample 2$\times$ & - & - & - & -\\
L13: Concat (L12, L4) & - & - & - & -\\
L14: Conv $3\times3$ & $128, 64$ & - & $1$ & Instance\\

L15: Conv $3\times3$ & $64, 32$ & - & $1$ & Instance\\
L16: Up-sample $2\times$ & - & - & - & -\\
L17: Concat (L16, L2) & - & - & - & -\\
L18: Conv $3\times3$ & $64, 32$ & - & $1$ & Instance\\
L19: Conv $1\times1$ & $32, 32$ & - & $1$ & Instance\\

L20: Reprojection\\
L21: MLP (Input 2) & $4, 16, 32$ & ELU & - & -\\
L22: Add (L21, L20) & - & - & - & -\\

\bottomrule
\end{tabular}
\vspace{-3mm}
\caption{Semantic feature extraction.}
\label{table:feature_extraction}
\end{table*}

\noindent \textbf{Cross-Reprojection Attention.} To model full semantic-aware dependencies from the 3D contextual space with computational efficiency, we design the Cross-Reprojection Attention module in Table~\ref{table:CRA_module} to learn dense and global contextual information, which can finally benefit the performance of semantic segmentation. The details of architecture and design can be found in Table~\ref{table:CRA_module} and Section 3.3 in the paper.

\begin{table*}[!h]
\centering

\begin{tabular}{lccc}
\toprule
Type & Feature dimension & Activation \\
\midrule
Input: Initial 3D contextual space & - & -\\
L1: Transpose (Input) & - & -\\
L2: Position Embeddings & - & -\\
L3: Add (L1, L2) & - & -\\
L4: Multi-head Attention (nhead=4) (L3) & 32 & ReLU \\
L5: Transpose (L4) & - & -\\
L6: Multi-head Attention (nhead=4) (L5) & 32 & ReLU \\
\bottomrule
\end{tabular}
\vspace{-3mm}
\caption{Cross-Reprojection Attention module.}
\label{table:CRA_module}
\end{table*}

\begin{table*}[!h]
\centering
\begin{tabular}{lccc}
\toprule
Type & Feature dimension & Activation \\
\midrule
Input 1: Initial 3D contextual space & - & -\\
Input 2: View direction differences & - & -\\
L1: Concat (Input 1, Input 2) & - & - \\
L2: MLP (L1) & $37, 16, 8, 1$ & ELU\\
L3: Sigmoid (L2) & - & -\\
\bottomrule
\end{tabular}
\vspace{-3mm}
\caption{Semantic-aware weight network.}
\label{table:semantic_aware_weight_net}
\end{table*}

\noindent \textbf{Semantic-aware weight network.} To construct the final semantic ray from refined 3D contextual space and learn the semantic consistency along the ray, we introduce the semantic-aware weight network in Table~\ref{table:semantic_aware_weight_net} to rescore the significance of each source view. More experiments about the semantic-aware weight net can be found in Table~\ref{tab:tau}, and we show architecture details in Table~\ref{table:semantic_aware_weight_net} and Section 3.4 of the paper.

\noindent \textbf{Geometry-aware network.} To build our generalizable semantic field, we adopt a geometry-aware network to predict density $\sigma$ and render the final semantic field with semantic logits. Moreover, we also leverage this network to produce the radiance and render a radiance field to show our rendering quality. We show the details of this network in Table~\ref{table:geometry_aware_net} and Section 3.4 of the paper.

\begin{table*}[!t]
\centering
\begin{tabular}{lccc}
\toprule
Type & Feature dimension & Activation\\
\midrule
Input: Initial 3D contextual space & - & -\\
L1: MLP (Input) & $32, 32$ & ELU\\
L2: MLP (Input) & $32, 1$ & ELU\\
L3: Sigmoid (L2) & - & -\\
L4: Dot-product (L1, L3) & - & -\\
L5: Cross-view Mean (L4) & - & - \\
L6: Cross-view Varience (L4) & - & -\\
L7: Concat (L5, L6) & - & - \\
L8: MLP (L7) & $64, 32, 16$ & ELU\\
L9: Multi-head Attention (nhead=4) (L8) & 16 & ReLU \\
L10: MLP (L9) & $16$ & ELU\\
L11: MLP (L10) & $1$ & ReLU\\
\bottomrule
\end{tabular}
\vspace{-3mm}
\caption{Geometry-aware network.}
\label{table:geometry_aware_net}
\end{table*}

% \begin{figure*}[!t]
%     \centering
%     \includegraphics[width=\linewidth]{cvpr2023/Figures/weight.pdf}
%     \caption{fddasfasd }
%     \label{fig:signifi}
% \end{figure*}

% \begin{figure*}[!t]
%     \centering
%     \includegraphics[width=\linewidth]{cvpr2023/Figures/semantic_render.pdf}
%     \caption{\textbf{Semantic rendering quality comparison}. On the left, we show direct semantic rendering results of our method and NeuRay~\cite{neuray} with semantic head. Limited by insufficient generalization, NeuRay with semantic head has difficulty to render semantics in unseen scenes and fails to capture contextual structure, while our method is able to learn the semantic structural prior, thus showing strong generalizability across different scenes. On the right, we show the experimental comparison between our S-Ray with $2k$ iterations finetuning ($\sim$10min) and Semantic-NeRF \cite{semantic-nerf} with $100k$ iterations. }
%     \label{fig:result}
% \end{figure*}

\begin{figure*}[!t]
    \centering
    \includegraphics[width=\linewidth]{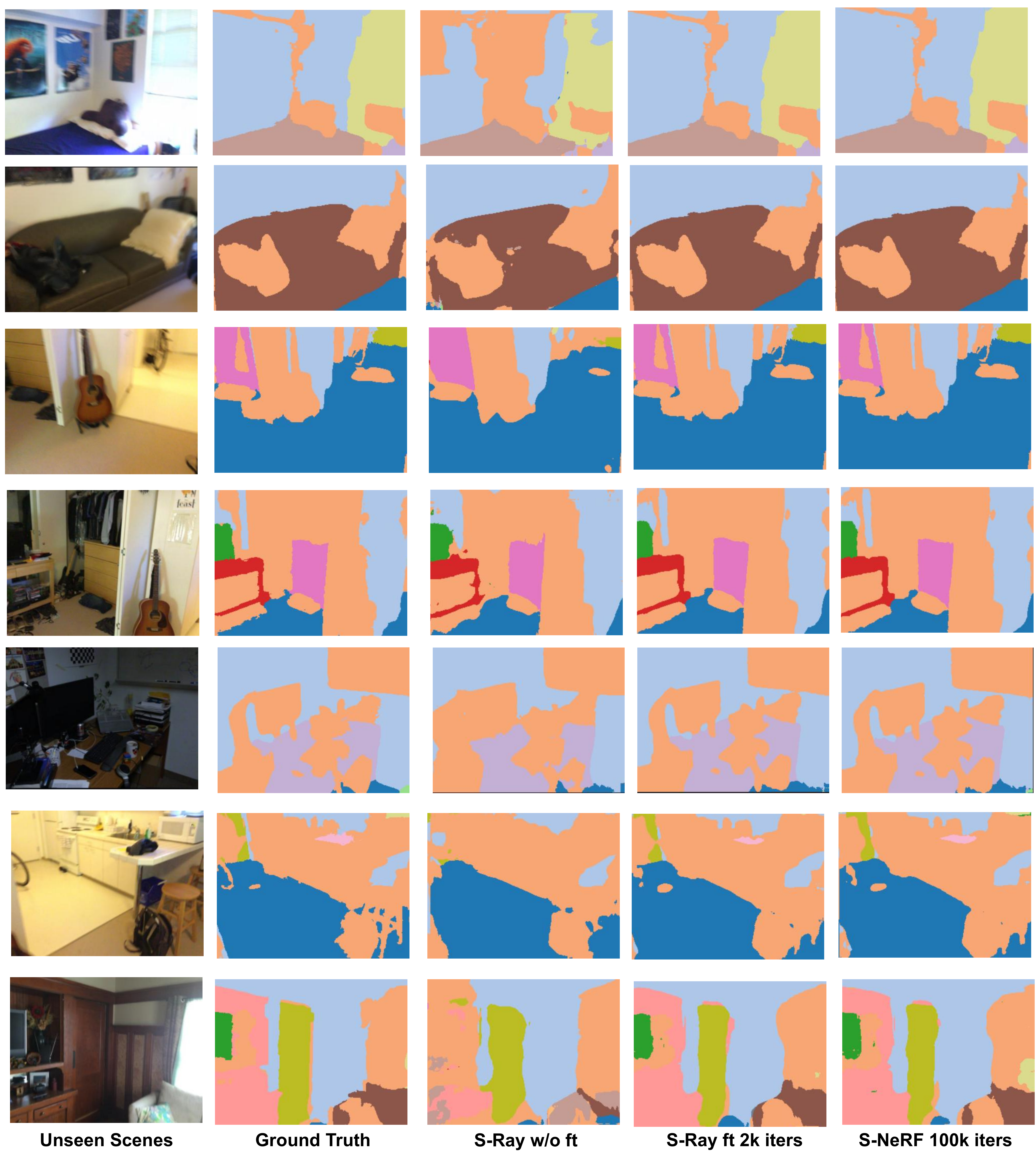}
    \caption{{Additional semantic rendering quality comparison}. More qualitative comparisons between our method S-Ray and non-generalizable method Semantic-NeRF~\cite{semantic-nerf} (S-NeRF for short) for semantic rendering in real data~\cite{scannet}. }
    \label{fig:more_semantic_result}
\end{figure*}

\begin{figure*}[!t]
    \centering
    \includegraphics[width=\linewidth]{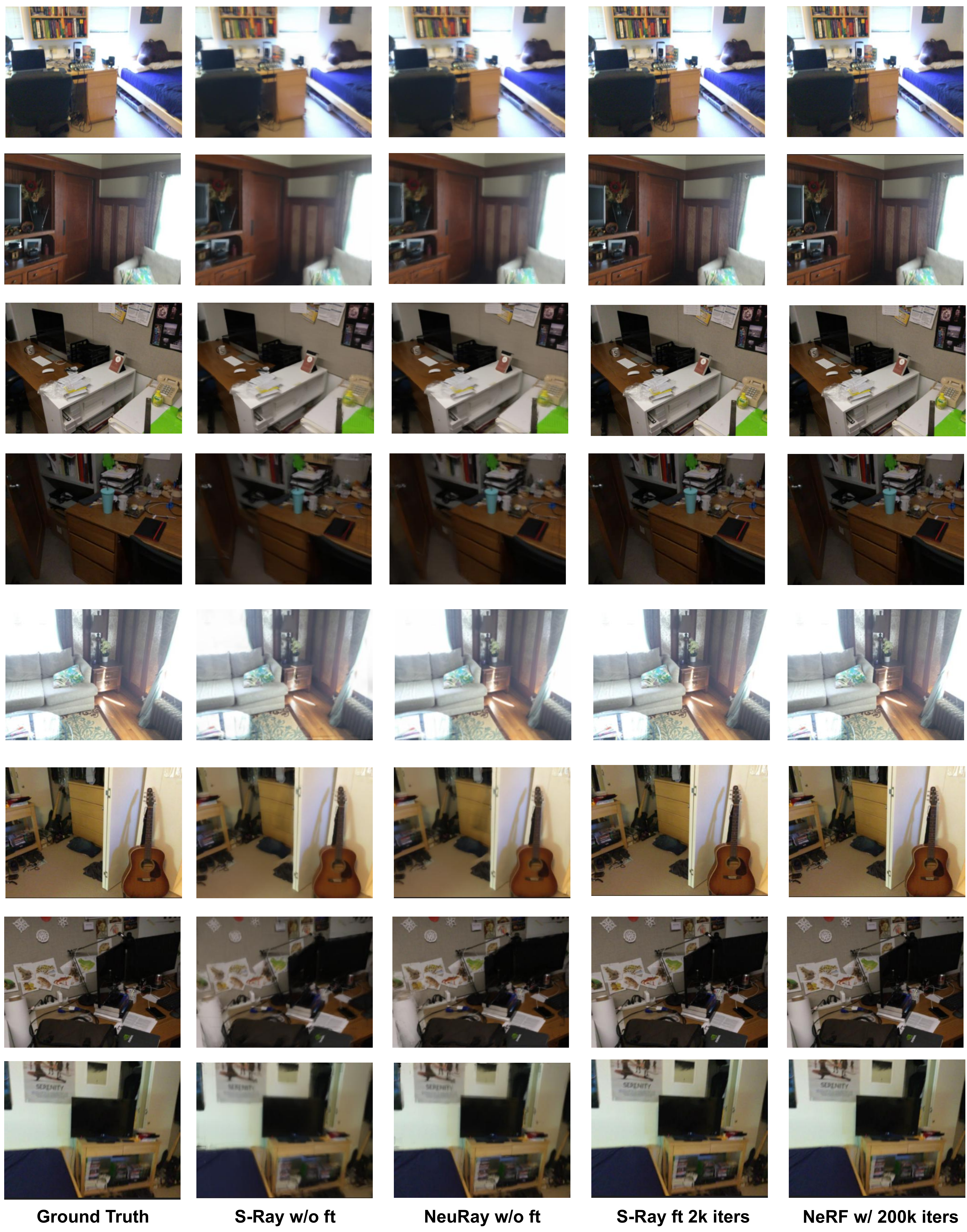}
    \caption{Qualitative results of scene rendering for generalization (w/o ft) and fine-tuning settings (ft) in real data~\cite{scannet}. Adding a color head from the geometry-aware network, We compare our method S-Ray with the generalizable rendering method NeuRay~\cite{neuray} and Valina NeRF~\cite{NeRF} with 200k iterations. }
    \label{fig:more_render}
\end{figure*}

\begin{table*}[!h]

\centering

\begin{tabular}{ccccccc}
\hline
\multirow{2}{*}{Method} & \multicolumn{3}{c}{Validation Set} & \multicolumn{3}{c}{Test Set} \\ \cline{2-7} 
                        & mIoU(\%)   & Total Acc(\%)    & Avg Acc(\%)   & mIoU(\%) & Total Acc(\%)  & Avg Acc(\%) \\ \hline
S-Ray (10k iters)       & 63.70   & 85.70        & 71.86     & 48.53 & 74.75      & 56.55   \\
S-Ray (50k iters)       & 72.85   & 88.72        & 79.52     & 52.32 & 77.31      & 59.38   \\
S-Ray (100k iters)      & 81.25   & 94.80        & 86.84     & 54.27 & 79.13      & 61.76   \\
S-Ray (200k iters)      & 89.31   & 97.91        & \textbf{91.10}     & 54.44 & 76.46      & 60.63   \\
\textbf{S-Ray (260k iters)} & \textbf{89.57} & \textbf{98.54} & 91.02 & \textbf{57.15} & \textbf{78.24} & \textbf{62.55}\\
S-Ray (300k iters)      & 88.99   & 98.40        & 90.39     & 55.84 & 77.67      & 62.15   \\ \hline
\end{tabular}%
\vspace{-3mm}
\caption{Quantitative results (mIoU, total accuracy, average accuracy) of our method (S-Ray) in training process from multiple scenes in real dataset~\cite{scannet}. }
\label{tab:training_detail}
\end{table*}

\begin{table*}[!h]
\centering
\begin{tabular}{cccccc}
\hline
Steps                 & Method        & mIoU(\%) &    Average Accuracy(\%) &    Total Accuracy(\%) & PSNR \\ \hline
\multirow{2}{*}{0}    & Ours          &  77.22   &        81.68          &      88.53         &   29.47   \\
                      & Semantic NeRF &   -   &              -    &            -    &  -    \\ \hline
\multirow{2}{*}{2k}   & Ours          &  92.66    &       98.73           &      98.73          &   29.80   \\
                      & Semantic NeRF &   78.32  &          82.69        &        85.81        &     20.62 \\ \hline
\multirow{2}{*}{4k}   & Ours          &   93.40   &        98.97          &       98.97         &   29.86   \\
                      & Semantic NeRF &   86.97   &        86.61          &        87.48        &    21.85  \\ \hline
\multirow{2}{*}{6k}   & Ours          &   94.17   &          99.06        &        99.06        &   29.92   \\
                      & Semantic NeRF &   87.08   &        87.85          &     88.01           &  22.62    \\ \hline
\multirow{2}{*}{8k}   & Ours          &  94.59    &        99.15          &      99.15          &    29.95  \\
                      & Semantic NeRF &  88.78    &        88.57          &        89.67        &   22.94   \\ \hline
\multirow{2}{*}{30k}  & Ours          &   95.10   &         99.43         &          99.42      &   30.05   \\
                      & Semantic NeRF &    91.78  &          94.86        &         95.78       &    24.78  \\ \hline
\multirow{2}{*}{100k} & Ours          &    -  &       -           &       -         &   -   \\
                      & Semantic NeRF &  95.05    &           98.73       &        99.02        &    29.97  \\ \hline
\end{tabular}%
\vspace{-3mm}
\caption{Performance of per-scene optimization in ScanNet~\cite{scannet}. We compare our method S-Ray with Semantic-NeRF~\cite{semantic-nerf} in per-scene optimization to show our fast generalizability in real data. Specifically, we choose the unseen scene0160\_02 for comparison. }
\label{tab:compare_semanticnerf}
\end{table*}

\end{document}